# ChatMOF: An Autonomous AI System for Predicting and Generating Metal-Organic Frameworks

*Yeonghun Kang, Jihan Kim\**

Department of Chemical and Biomolecular Engineering, Korea Advanced Institute of Science and Technology (KAIST), 291, Daehak-ro, Yuseong-gu, Daejeon 34141, Republic of Korea

# ABSTRACT

ChatMOF is an autonomous Artificial Intelligence (AI) system that is built to predict and generate metal-organic frameworks (MOFs). By leveraging a large-scale language model (GPT-4 and GPT-3.5-turbo), ChatMOF extracts key details from textual inputs and delivers appropriate responses, thus eliminating the necessity for rigid structured queries. The system is comprised of three core components (i.e. an agent, a toolkit, and an evaluator) and it forms a robust pipeline that manages a variety of tasks, including data retrieval, property prediction, and structure generations. The study further explores the merits and constraints of using large language models (LLMs) AI system in material sciences using and showcases its transformative potential for future advancements.

# Introduction

The realm of generative artificial intelligence (AI) is witnessing an unprecedented surge, predominantly fostered by a new generation of computational tools known as large-scale language models (LLMs)[1-5]. These innovative models are deeply rooted in a novel architectural design paradigm, referred to as transformer models[6]. Their capabilities, however, stretch far beyond the domain of basic language tasks. Emulating human thought processes, they exhibit logical functions such as learning from a limited number of examples (few-shot learning), or in certain situations, learning without any examples (zero-shot learning).[3,5,7] This is achieved through the understanding of vast volumes of text data, underscoring the immense potential held by these models. A significant development in this rapidly advancing field is the rise of autonomous LLM agents, which uses LLMs for a wide range of tasks[8-10]. This system, powered by multiple LLMs, excels in making decisions on its own. As such, the idea of using LLMs to independently reason, find information, and produce answers is becoming more popular across different areas of research fields[11-14].

Despite marked progress in application of LLM across diverse fields[15,16], the full potential of its advanced technology within materials science remains largely untapped. This limitation primarily stems from two considerable challenges. Firstly, the inherent complexity of materials poses a significant hurdle as these materials often lack sufficient text-compatible input representations, thereby failing to fully encapsulate their complete properties[17-20]. This difficulty in encoding materials for LLMs restricts their understanding and processing capabilities. Secondly, there is a notable scarcity of material-specific training data in the field. In comparison to other disciplines, materials science lags behind due to fewer dedicated databases and their associated data, exacerbating the challenge of representing this scant data in a text format suitable for LLMs.

Despite these obstacles, there are ongoing attempts to leverage the capabilities of LLMs in materials science.[21-23] However, so far, these efforts have primarily focused on extracting data from scientific literature and generating responses based on this extracted data, with the actual material itself remaining a largely untouched resource. As such, the exploration and realization of the full potential of LLMs within the sphere of materials science still beckons.

In this work, we highlight the development of an innovative methodology that utilizes autonomous AI to generate new materials and to predict properties of the materials with the focus specifically on metal-organic frameworks (MOFs)[24-26]. MOFs are used in many chemical applications[27-30] due to their large porosity[31-33], high surface area[32], and exceptional tunability[34]. To this end, we have developed the autonomous AI system for MOF (called ChatMOF), which holds the potential to predict MOF properties from text-based inquiries and to generate MOFs with specified properties (i.e. inverse design). This pioneering approach can potentially significantly bridge the gap between the novice users and the computational and machine learning tools, which can potentially facilitate the progress in developing new materials for various applications.

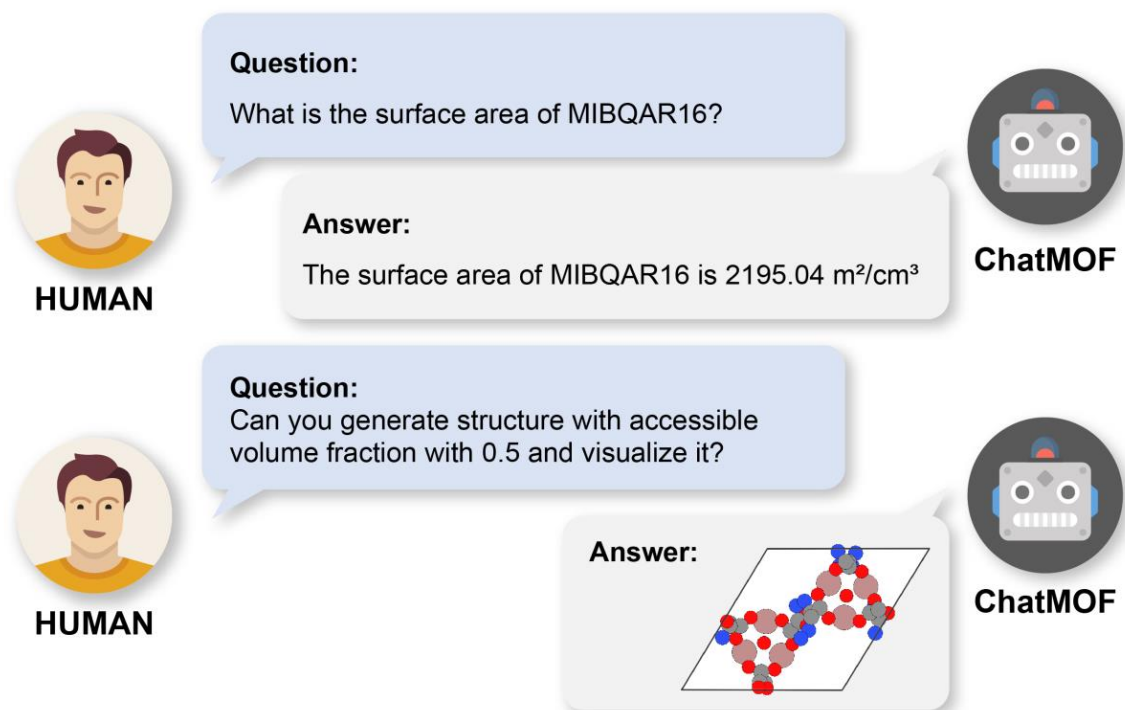

**Figure 1.** A Conceptual image that explains the ChatMOF. When a user poses a textual question about the properties of a MOF, an appropriate answer is provided by ChatMOF. If a user desires to generate a new MOF, ChatMOF is capable of creating a new MOF that satisfies the condition.

# Results and Discussion

## Design for ChatMOF

The effectiveness of autonomous LLM agents is predicated on its capability to accurately extract essential details from textual inputs and offer relevant responses, irrespective of the presence of a rigidly structured query.[21] This concept is vividly illustrated in ChatMOF, as demonstrated in Figure 1. A user may pose a query in textual form regarding the properties of a material, to which ChatMOF responds by supplying a detailed description related to the material in question. Moreover, the operational scope of this system extends beyond the simple retrieval of information. When a user expresses the need to generate a MOF with specific properties, ChatMOF is capable of generating the requested material structure accordingly.

In the context of ChatMOF, LLMs performs the role of a central planner, regulator and evaluator, akin to a human brain. Although LLM effectively fulfills the role of reasoning, its proficiency in specialized tasks falls short. While a single LLM's proficiency in specialized tasks may be limited[15], LLMs offer an effective way of integrating and employing various databases and machine learning models seamlessly due to their inherent capacity for reasoning[35,36]. ChatMOF utilizes the LLM to orchestrate a comprehensive plan and employ toolkits for information gathering, comparable to an expert devising a strategy, writing the code, and executing the plan. This synergy allows the system to precisely predict material properties, deduce synthesis techniques, and fabricate new materials with preset properties.

As depicted in Figure 2, ChatMOF is composed of three main components: an agent, toolkit, and an evaluator. The agent responds to human queries by traversing four primary steps (i.e. thoughts, actions, action inputs, and observation), mirroring the methodology outlined in the ReAct[37] and MRKL papers[38]. Initially, the user's query is set as the end goal, followed by contemplation about the path to achieve this goal. Subsequently, ChatMOF decides on the appropriate toolkit to employ from the available options. After the chosen toolkit is executed, the observed results serve as the basis for evaluating whether a final answer can be generated. If feasible, the final answer is presented, otherwise, the process cycles back to the thought step to formulate a new strategy.

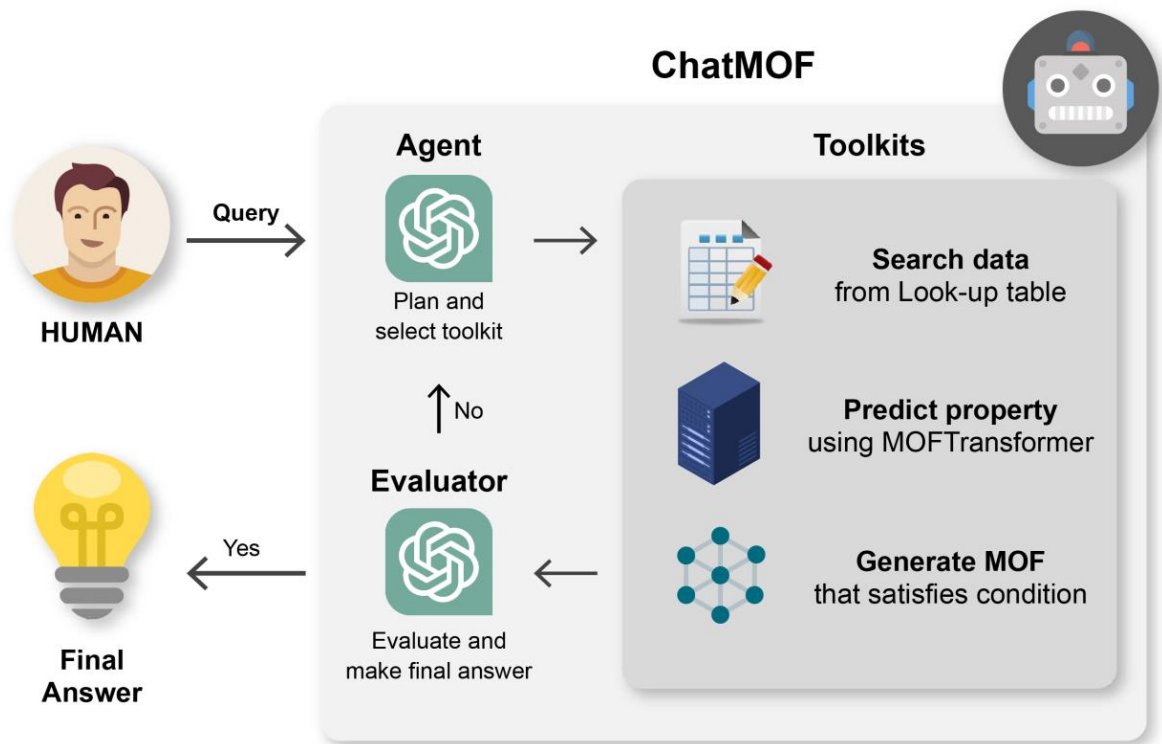

**Figure 2.** The schematic image of ChatMOF. ChatMOF comprises three core components: an agent, toolkits, and an evaluator. Upon receiving a query from human, the agent formulates a plan and selects a suitable toolkit. Subsequently, the toolkit generates outputs following the proposed plan, and the evaluator makes these results into a final response.

## Toolkits

ChatMOF employs an assortment of toolkits to acquire, predict, or generate material information. These toolkits can be primarily classified into five categories: table-searcher, internet-searcher, predictor, generator, and others. Table-searcher involves obtaining desired information from existing data, while internet-searcher employs internet resources, such as Google and Bing, to produce desired information. The predictor utilizes machine learning models to obtain specified properties. The generator refers to the toolkit that constructs material structures fulfilling certain properties. Lastly, the others encompass a variety of aids like calculators, file saving and reading functions, and visualization toolkits.

Due to the facile synthesis MOF structures, there are many different database associated with the MOF structures: (1) computational-ready experimental MOFs (CoREMOF)[39,40] and (2) quantum MOF (QMOF) database[41]. The CoREMOF database is an archive of synthesized materials present in a CSD MOF subset[42], encompassing computations of various properties of MOFs including geometric descriptors. The QMOF database is populated with electrical property data, such as bandgap, formation energy, homo, and lumo, derived through DFT calculations. When a user demands these electrical properties, ChatMOF seeks and extracts them from the QMOF database.

As such, if a user seeks information about a specific MOF that is included in these databases, ChatMOF can locate and deliver the desired information from the pre-tabulated data. Figure 3 provides an illustrative example of a table-search operation conducted by ChatMOF. When faced with a user query, ChatMOF devises a strategic approach to extract the requisite data. Subsequently, it creates a Python code tailored to retrieve specific information from the database in accordance with the premeditated strategy. This code typically uses the "pandas[43]" library to extract or filter relevant details. The drafted code is then executed within the ChatMOF's designated executor. Upon observing the outcome of this operation, ChatMOF contemplates the necessary steps to derive the final answer, culminating in the delivery of the requested response.

The versatility of ChatMOF extends to handling diverse table data derived from text mining or rule-based coding processes. For questions related to the building blocks of a MOF, the MOFkey[17] database proves to be instrumental. This particular database leverages rule-based methods to obtain insights about the organic linkers and metal clusters of a MOF, providing details about its topology and the potential presence or absence of interpenetration. In addition, for users seeking guidance on MOF synthesis, the DigiMOF[44] database becomes a valuable resource. DigiMOF provides an array of synthesis conditions, extracted via text mining techniques from MOF-related academic papers, and includes information on organic and metal precursors, and solvent.

**Question: How high is the accessible surface area of JUKPAI?**

Thought: I need to find the accessible surface area of JUKPAI
Action: search_csv
Action Input: "Search name JUKPAI and provide information of its accessible surface area"

[Table Searcher] Thought: To search for a specific value in the DataFrame, we can use the ...
[Table Searcher] Input:

```python
print (df.loc[df['name'] == 'JUKPAI',
        'Accessible Surface Area (m^2/cm^3)'].to_markdown())
```

[Table Searcher] Observation:

|      | Accessible Surface Area (m^2/cm^3) |
|------|------------------------------------|
| 4837 | 1474.22                            |

[Table Searcher] Final Thought: The accessible surface area of name 'JUKPAI' is 1474.22 ...

Thought: I now know the final answer

**Answer : The accessible surface area for JUKPAI is 1474.22 m²/cm³**

**Figure 3.** Example of a Table Searcher for the question "How high is the surface area of JUKPAI?". The human's question prompts the system to devise a strategy. From this, Python code is generated and executed to extract the desired information, which is then returned as the final answer.

The accuracy of the look-up table search is contingent on the pre-calculated values available in the specific files. And for queries regarding the properties of MOFs that are not available, computational simulation can serve as an attractive alternative method, but unfortunately, simulations are a time-intensive process and an abundance of computational resources[45]. The best resolution to such challenges is the application of machine learning models, which enable high-accuracy predictions grounded in extensive data. In the case of pre-trained machine learning models, predictions are quick and can be made for a significant volume of substances simultaneously, making it an excellent toolkit for integration into ChatMOF.

As an appropriate toolkit for the prediction task, ChatMOF uses the MOFTransformer[46,47] model that has been developed in our group for the universal prediction of MOF properties. This model leverages both local features, such as atoms and bonds, and global features like surface area and topology. Having undergone pre-training with one million hypothetical MOFs and subsequent fine-tuning for specific properties, MOFTransformer shows high performance in predicting various properties. Moreover, it affords the convenience of predicting the desired property by retaining the model structure while altering the weights of a model that's been fine-tuned for the desired property.

The key to generating accurate responses is selecting the appropriate fine-tuned model with the MOFTransformer and the material to which it will be applied, based on the query. Similar to HuggingGPT[10], ChatMOF does not directly compute the material properties, but it rather selects the appropriate machine learning model. Figure 4 shows the example prompt for ChatMOF using the MOFTransformer predictor. When a user asks a question such as "At room temperature (298K), what's the CO2 Henry coefficient for XEGKUR?", generator develops a strategy for solving the problem. The optimal fine-tuned model of MOFTransformer that aligns with the plan, in this case,

'CO2_henry_coefficient_298K' is chosen, and the material to predict (XEGKUR) is decided. Following this, the selected model is employed to make prediction. Finally, it make a response to the query using a language learning model based on the observed values from the machine learning model. Since this is a logarithmic value, ChatMOF selects the "Python_REPL" tool and calculates an exponential value to get the final answer.

**Question: At room temperature (298K), what's the CO₂ Henry coefficient for XEGKUR?**

Thought: I need to find the CO₂ Henry coefficient for XEGKUR at room temperature
Action: search_csv
...
Thought: The search_csv tool did not provide any information on the CO₂ Henry coefficient...
Action: predictor
Action Input: Predict the CO₂ Henry coefficient for XEGKUR at 298K

[predictor] Thought: I need to gather data on XEGKUR and train a machine learning to pre...
[predictor] Property: CO2_henry_coefficient_298K
[predictor] Materials: XEGKUR

```bash
load model : CO2_henry_coefficient_298K/best.ckpt
Predicting DataLoader 0: 100% |■■■■■■■■■■■■■| 1/1  [00:00<00:00]
```

[predictor] Final Thought: Based on the results, answer the question using the predicted ...

Observation: The predicted CO2 Henry coefficient at 298K for XEGKUR_clean is **-3.62769 mol/Kg·Pa**. However, this is a **logarithmic value**. To get the original value, an exponential must be applied.
Thought: I need to convert the logarithmic value to the original value
Action: Python_REPL
Action Input:

```python
import math
print (math.exp(-3.62769))
```

Observation: 0.026577507595890823
Thought: I now know the final answer

**Answer : The CO2 Henry coefficient for XEGKUR at room temperature (298K) approximately 0.027 mol/Kg·Pa**

**Figure 4.** Example of a predictor for the question "At room temperature (298K), what's the CO2 Henry coefficient for XEGKUR?" The predictor sets up a plan to solve the question, an appropriate model, and target material. Based on this, it uses machine learning to predict the value, which is then used to derive the final answer.

Finally, a key aspiration among researchers in the field of MOFs is the inverse design of MOFs exhibiting desired properties. In materials science, various generative models, including Generative Adversarial Networks (GAN)[48,49] and Diffusion models[50], have been employed for inverse design. However, due to the inherent complexity of MOFs, which includes a large number of atoms, large void fraction, and complex topologies, an atom-by-atom inverse design approach has been elusive. As a workaround, MOF inverse design has been facilitated top-down approaches leveraging through genetic algorithms[51-53], Variational Autoencoders[54] (VAE), or reinforcement learning[55] for the selection of building blocks and their placement into suitable topologies.

Genetic algorithms are notably suitable for integration with LLMs. As a bio-inspired optimization methodology, genetic algorithms operate on a selection, mutation, and crossover principle, making them adaptable and efficient.[56] For their application to MOFs, these frameworks must be delineated by genes comprising topology and building blocks. For instance, a representative MOF, HKUST-1, can be depicted as tbo+N17+N10, with tbo representing topology and N17 and N10 representing the building block notations. As these gene representations are textual, they facilitate the application of genetic algorithms using an LLM.

Figure 5 showcases the utilization of a genetic algorithm by ChatMOF to fabricate a MOF per user specifications. Upon a user query, the system develops a strategy optimized for the genetic algorithm. It also identifies the target property and determines the loss function most suited for the objective, such as choosing the maximum, minimum, or closest value. Guided by this plan, ChatMOF selects parent genes from the existing database that align with the loss function. Using LLM, it produces children genes from the selected parent genes that are most likely to achieve the target. These children are then transformed into a structure file, and their properties are estimated through machine learning. This procedure is reiterated a fixed number of times, generating multiple generations of children with each generation yielding MOFs progressively nearer to the target. From the created structures, the one that aligns most closely with the question is finally chosen and presented as the response.

**Question: Can you generate the structures with the largest surface area?**

Thought: I need to generate structures with the largest surface area
Action: generator
Action Input: "Generate structures with the largest surface area"

[generator] Thought: I need to generate structures with the largest surface area
[generator] Objective: max
[generator] Predict Property: accessible_surface_area

```bash
load model : accessible_surface_area/best.ckpt
Predicting DataLoader 0: 100% | ■■■■■■■■■■■■■■■| 1/1 [00:00<00:00]
```

[generator] Find Parents: extract **cif_id** and **accessible_surface_area** of 100 materials ...
... *(search table)*

|       | cif_id       | Accessible Surface Area (m^2/cm^3) |
|-------|--------------|-------------------------------------|
| 2664  | pcu+N47+E33  | 5808.59                             |
| 1411  | pcu+N274+E32 | 5714.84                             |
| 8     | pcu+N613+E90 | 5665.73                             |
| ...   | ...          | ...                                 |

[generator] Get Children: generate 100 new materials with the largest surface area from ...
[generator] Generate Structures:

```bash
Generate cifs: 100% | ■■■■■■■■■■■■■■| 1/1 [00:00<00:00]
```

... *(run N cycles)*
[generator] Final Thought: Based on the generated CIF, find the materials that has the ...
... *(search table)*

|                        | **12718**          |
|------------------------|--------------------|
| cif_id                 | rtl+N535+N234      |
| accessible_surface_area | 6411.28173828125  |

Observation: The answer for question is rtl+N535+N234
Thought: I now know the final answer

**Answer : The structure with the largest surface area is rtl+N535+N234**

**Figure 5.** Example of a generator for the question "Can you generate the structures with the largest surface area". The generator establishes a plan, objective and property for the human question. Based on this, it finds parents that satisfy the objective. It uses a genetic algorithm to create children genes and generate structures. This is repeated for a number of cycles to generate new MOFs, which are used to derive the final answer.

Moreover, ChatMOF is engineered to perform a diverse set of toolkits, which extend beyond the realms of LLMs. This includes capabilities such as file search, Internet search, and even simple calculations. These additional functionalities are primarily enabled by leveraging the varied capabilities provided by LangChain[57], enhancing the overall functionality and utility of ChatMOF. Thus, it is not merely a material analysis tool, but a comprehensive system that can accommodate a wide array of tasks and operations.

In addition, ChatMOF incorporates the Atomic Simulation Environment (ASE)[58] library as an integral toolkit to facilitate diverse operations on material structure data. The ASE library holds considerable importance in the field of materials science due to its capabilities, including atom manipulation, cell information acquisition, and visualization, among others. Similar to the function of a table searcher, when confronted with a query, ChatMOF devises a strategic plan and constructs suitable Python code utilizing the ASE library to fulfil the query's demands. Subsequently, this code is executed.

## Evaluation

To evaluate performance of ChatMOF, analysis was conducted for "search task", "prediction task", and "generation task". For evaluation purposes, questions for ChatMOF were created utilizing GPT-4.0 to generate various sentences about the given properties of a MOF. The respective questions for each task can be found in Table S1-3. Accuracy, gauging how adequately the logic responded to the question, was measured for each task. An analysis of the accuracy of ChatMOF utilized three labels: "True", "False (token limit exceeded)", and "False (logic error)". The label "True" signifies that ChatMOF's logic was precise and the yielded answer was accurate. The term "False (Token Limit Exceeded)" was used when the token count in LLM surpassed the maximum allowance of 4,000, thus obstructing further progress. Lastly, the "False (Logic Error)" label designated situations where an error in ChatMOF's logic resulted in an incorrect response or an anomaly. Such situations typically occur when an erroneous plan for obtaining an answer was devised or when an error in output interpretation diverts the system from the desired direction.

Figure 6 presents the accuracy measurements for the three tasks using ChatMOF with GPT-4. Accuracy was measured for 100 sample questions for the search and prediction tasks, and 10 sample questions for the generation task. The number in the bar graph indicates the number of each question in each class. Both the search and prediction tasks rendered accurate answers with high frequency. Excluding 'Token Limit Exceeded' instances (4 out of 100, 6 out of 100, and 2 out of 100, for search, prediction, and generation tasks respectively), they exhibit high accuracies of 96.9% and 95.7%, respectively. For the generation task, the accuracy stood at 87.5%. Given its complexity, relative to the other two tasks, the accuracy rate is lower. Regardless, all three tasks report high accuracy rates, and these tasks carry significance weight because these are tasks that a typical LLM fails to execute flawlessly. A single LLM fails to provide precise information since it lacks specific information about the material, particularly for obtaining property information, which is challenging to source via an internet search.

Also ChatMOF, when integrated with GPT-4, exhibits superior performance compared to its integration with GPT-3.5-turbo. As evidenced in Figure S2, the accuracy of ChatMOF with GPT-3.5-turbo stands at 95%, 91%, and 77.8% for the search, prediction, and generation tasks respectively, excluding instances of "Token Limit Exceeded". Across all tasks, GPT-4 consistently outperforms GPT-3.5-turbo in accuracy. This enhanced accuracy of GPT-4 can be attributed to its refined reasoning and comprehension capabilities, particularly during the planning phase. Figure S3 illustrates the distinct approaches that GPT-4 and GPT-3.5-turbo take when presented with the same query: "How does the pore limiting diameter of YUSGID_clean compare with other materials?". While GPT-3.5-turbo seeks the values for all materials mentioned in the query, leading to a token error and subsequent inability to provide an answer, GPT-4 adopts a more holistic strategy. It assesses the distribution of all materials, leveraging metrics such as mean, variance, and quartile values of the property in question. This approach enables GPT-4 to determine the relative position of the target material in the overall distribution, thus delivering a more informative response to the user.

For the "search task," the writing of code utilizing the pandas library significantly impacts the accuracy. 'Token Limit Exceeded' generally occurs when the output code surpasses the permissible token count. This frequently arises when all relevant materials that satisfy a given condition are provided (for example, when a list of materials with a particular property is listed), or when the question contains a comparative clause such as "compared to other materials." 'Logic Error' typically surfaces when there is a flawed strategic approach or a code error. An instance of this would be when a request to provide 10 specific items is met with a misguided strategy that solely aims to "extract high values," failing to retrieve the specified number of items.

During the "prediction task," difficulties often occur in the interpretation process of the observed values using machine learning techniques. Both the 'Token Limit Exceeded' and 'Logic Error' occurrences can stem from the effort to draw the correct answer from the table based on the estimated values. 'Logic Errors' can manifest not only during the table search phase but also during the strategy formulation stage. An erroneous strategy could either lead to the loading of an unsuitable model or to the generation of an input that is incompatible with the intended model.

The "generation task" presents a more intricate set of challenges, inviting a variety of errors. A frequently observed 'Logic Error' appears when no parent genes can be retrieved from database. If the objective function aims for maximum or minimum values, a satisfying parent gene can always be found. However, if the goal is to get close to a certain value or to fall within a specific range, the selected range might not yield any satisfying parent genes. In such scenarios, the strategy is adapted to incorporate more data. However, if no suitable parent genes are found even after modifying the strategy, it results in an error. Further, both 'Token Limit Exceeded' and 'Logic Error' might occur during the extraction of the most suitable MOF from the generated MOFs, aligning with the objective function.

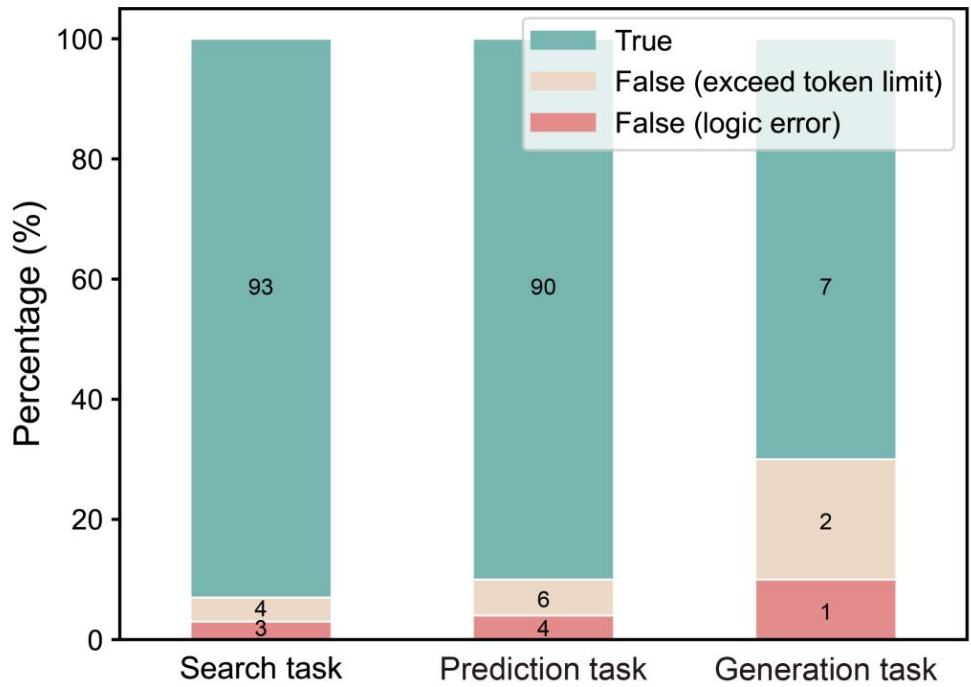

**Figure 6.** Depiction of accuracies for three tasks using GPT-4 model - search, prediction, and generation. Accuracies were evaluated based on three labels: True, False (exceeding token limit), and False (logical error). The number in the bar represent the count of each label.

## Inverse Design Validation

One notable observation is that with each generation, the genetic algorithm refines the distribution of material properties to better align with the target value. Figure 7 illustrates the outcomes of the generation task for two different scenarios.

Figure 7(a) reveals the structures generated in response to the question, "Can you generate structures with the largest surface area?" In this case, ChatMOF interpretes the property as accessible surface area, with the objective as maximizing this parameter. The initial generation MOF (0th generation) displays a broad distribution of surface area with an average value of 3,748 $m^2/g$. However, with each subsequent generation, the peak at a higher position amplifies. By the third generation, the offspring MOF exhibits a significantly elevated average peak value of 5,554 $m^2/g$.

Analogously, Figure 7(b) portrays the distribution of structures generated to fulfill the request "I'm looking for structures with a hydrogen uptake of about 500 $cm^3/cm^3$ at 100 bar and 77 K, can you generate those?". Here, ChatMOF sets the property to hydrogen uptake at 100bar and 77 K with the objective of achieving close proximity to 500 $cm^3/cm^3$. The distribution of the initial structure spans evenly from 250 $cm^3/cm^3$ to 650 $cm^3/cm^3$. However, the structure created in the final generation displays the most pronounced and narrow peak at 500 $cm^3/cm^3$. This indicates the efficiency of the genetic algorithm utilizing the LLMs.

Figures 7(c) and 7(d) depict the final structures for the queries in 7(a) and 7(b). The optimal structure in 7(c), rtl+N535+N234, boasts the highest surface area amongst the generated MOFs. The predicted value stands at 6411.28 $m^2/g$. Upon performing a geometric optimization and calculating accessible surface area using Zeo++[59], the surface area is revealed to have 7647.62 $m^2/g$. This value is notably higher when compared to the CoREMOF database. Figure S1 illustrates the distribution of accessible surface areas within CoREMOF. This particular structure's surface area ranks the third-highest position in the CoREMOF ranking. In a similar vein, the optimal configuration of dia+N719+E186, showcased in Figure 7(d), possesses a surface area of 499.998 $cm^3/cm^3$, mirroring the stipulated target of 500 $cm^3/cm^3$. Following geometric optimization of this structure, its uptake was calculated using RASPA, yielding a value strikingly close to the goal, at 495.823 $cm^3/cm^3$.

Despite its successes, the generation task of ChatMOF does present some limitations. Chief among these is the decrease in gene diversity due to constraints on input and output tokens. The token count restricts the number of parent and child structures to around 100, a fraction compared to inversed design studies that employ conventional genetic algorithm procedures that generate upwards of 100,000 structures for each generation. Other constraints, such as the limited number of topologies and cycles, stem from resource and time restrictions. Yet, despite these limitations, ChatMOF excels in generating MOFs fitting the objective function, attesting to its efficacy.

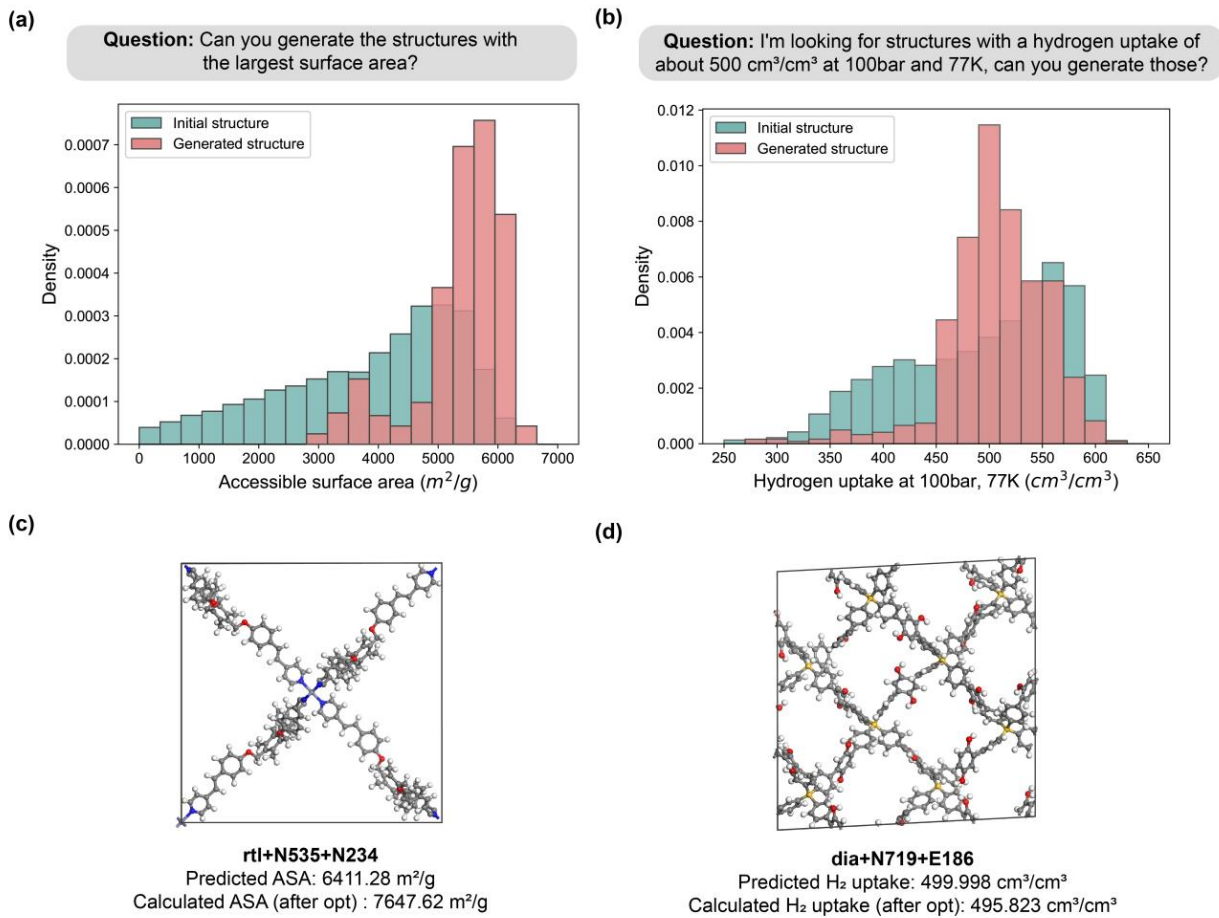

**Figure 7.** (a) Histogram depicting the initial structure and the generated structure for a question concerning the maximum value of surface area. (b) Histogram showing the initial structure and the generated structure for a query where the value of hydrogen uptake is set close to 500. (c) Illustration of the MOF with the largest surface area as generated by ChatMOF. ASA stand for accessible surface area. (d) Representation of the MOF with an $H_2$ uptake value closest to 500 $cm^3/cm^3$ at 298K, 1bar, as generated by ChatMOF.

## Collaborative Online Platforms

One limiting factor of ChatMOF is the performance reliance on the number of pre-trained weights in the MOFTransformer used in the predictor task. An increased quantity of fine-tuned weights allows for the prediction of more properties, thereby enabling more active prediction and generation processes. However, each user faces constraints on the number of models that can be utilized, given that it is unrealistic for one individual to possess all the data.

To train a model, the collection of experimental data or the execution of computational simulations is necessary. While some calculations, such as pore limiting diameter or surface area, demand less time, other tasks such as band-gap, homo, and lumo calculations are considerably more computationally demanding. The generation and training of data for these complex tasks can be quite cumbersome.

To address these issues, there is a need to create an online platform akin to HuggingFace, where users can freely post their learned weights. An example of this model would be HuggingGPT[10], which functions by selecting the most appropriate model among those posted on HuggingFace. Should users upload their trained models built on data they have uploaded onto this platform, it will enable other users to access them. Upon the posting of new weights online, ChatMOF will review them and if the required data is available online, the model will be downloaded automatically. The existence of this online platform will reinforce ChatMOF as a potent toolkit for predicting MOF properties. Moreover, pre-calculated data, such as those from multiple mining, can also be employed for table searches. If data sharing is executed effectively, superior results can be achieved collectively.

## Conclusion

The investigation into the role of generative AI in materials science, specifically through the lens of ChatMOF, unveils substantial potential for predicting and generating MOFs. This unique system, which bridges the transformative capabilities of AI and the intricate facets of materials science, demonstrates exceptional performance across various tasks. The accuracy analysis reports high success rates, notably 96.9% and 95.7% for the search and prediction tasks, respectively. Meanwhile, the more complex structure generation task, despite its intricacy, yields a notable accuracy rate of 87.5%. These promising results underline the efficacy of ChatMOF, even when confronted with the most demanding tasks. Despite certain limitations, such as dependence on the number of pre-trained weights, ChatMOF symbolizes a significant stride towards fully autonomous AI in the realm of materials science. As the technology evolves, and with a systematic enhancement of the model's capacity and data sharing across an online platform, ChatMOF's performance could be further optimized, paving the way for unprecedented advancements in MOF research.

# Method

ChatMOF operates via the LangChain[57] library. LangChain serves as a platform for accessing diverse modules within a Large Language Model (LLM), streamlining prompt engineering in an LLM environment. ChatMOF integrates various toolkits from LangChain alongside its distinct toolkit. For the roles of agent, evaluator, and toolkit within ChatMOF, OpenAI's Chatmodel, GPT-4, and GPT-3.5-turbo LLMs are employed. During the experiments, the temperature parameter was calibrated to 0.1.

The searcher component of ChatMOF adopts the CoreMOF structure, enriched by geometric features derived through ZEO++[59]. In instances of code discrepancies, corrections are made up to a threshold of three attempts. The predictor module within ChatMOF leans on MOFTransformer, trained on insights from four academic articles. Notably, MOFTransformer operates under version 2.1.2.

The generative aspect of ChatMOF is structured around three iterative cycles. This generator employs a genetic algorithm across nine unique topologies, namely pcu, dia, acs, rtl, cds, srs, ths, bcu, and fsc. For every topology, a batch of 100 offspring genes arises from a set of 100 parental genes, chosen from a foundational group of 2000 MOFs. Structures are then formulated based on these newly minted genes, followed by value computation via the predictor. This cycle refines the pool of parental genes, and after the designated cycles, an optimized target structure is procured from the cumulative data.

## Conflicts of interest

There are no conflicts to declare.

## Author Contributions

Y.K developed ChatMOF and wrote the manuscript with J.K. The manuscript was written through the contributions of all authors. All authors have given approval for the final version of the manuscript.

## Code availability

The ChatMOF library is available at https://github.com/Yeonghun1675/ChatMOF.git.

## Acknowledgements

Y. K., and J. K. acknowledge funding from the National Research Foundation of Korea (NRF) under Project Number 2021M3A7C208974513 and 2021R1A2C2003583. This work was supported by the National Supercomputing Center with supercomputing resources including technical support (KSC-2022-CRE-0515).

# ChatMOF: An Autonomous AI System for Predicting and Generating Metal-Organic Frameworks

*Yeonghun Kang, Jihan Kim\**

Department of Chemical and Biomolecular Engineering, Korea Advanced Institute of Science and Technology (KAIST), 291, Daehak-ro, Yuseong-gu, Daejeon 34141, Republic of Korea

# Table of Contents

**Supplementary Note S1.** Prompt for planning in the searcher

---

You are working with a pandas dataframe in Python. The name of the dataframe is `df`.

You should make a valid python command as input. You must use print the output using the `print` function at the end.

You should use the `to_markdown` function when you print a pandas object.

Use the following format:

Question: the input question you must answer

Thought: you should always think about what to do

Input: the valid python code only using the Pandas library

Observation: the result of python code

... (this Thought/Input/Observation can repeat N times)

Final Thought: you should think about how to answer the question based on your observation

Final Answer: the final answer to the original input question. If you can't answer the question, say `nothing`

The index of the dataframe must be be one of {df_index}. If it's not in the index you want, skip straight to Final Thought.

{information}

Begin!

Question: What is the head of df? If you extracted successfully, derive 'success' as the final answer

Thought: To get the head of a DataFrame, we can use the pandas function head(), which will return the first N rows. By default, it returns the first 5 rows.

Input:
```
import pandas as pd
import json
print(df.head().to_markdown())
```

Observation: {df_head}

Final Thought: The head() function in pandas provides the first 5 rows of the DataFrame.

---

Final Answer: success

Question: {question}
{agent_scratchpad}

**Supplementary Note S2.** Prompt for planning in the predictor

plan to use machine learning to predict the properties of matter. To answer the question, you have to fill in the following format:

Question: the input question you must answer

Thought: you should always think about what to do

Property: the property you can predict, should be one of [{model_names}]

Material: names of materials separated using comma. If you need to proceed for all material, write *. To proceed for a specific topology, append the topology name with an * (ex. pcu*)

... (this Property/Material can repeat N times)

Run Machine Learning: nothing to do

Final Thought: you should think about how you will derive a final answer from the results of machine learning.

Begin!

Question: predict the surface area and hydrogen uptake of ACOPEN and ELOBPE.

Thought: I need to gather data on ACOPEN and train a machine learning model to predict its surface area and hydrogen uptake.

Property: surface_area

Material: ACOPEN, ELOBPE

Property: hydrogen_uptake_100bar_77K

Material: ACOPEN, ELOBPE

Final Thought: Based on the result, answer the question using predicted surface area and the predicted hydrogen uptake at 100 bar.

Question: which MOF has a highest band-gap?

Thought: I need to gather data on the band-gap of different structures and compare them to find the one with the highest value.

Property: bandgap

Material: *

Final Thought: Based on the result, find the structure with the highest predicted band-gap value.

Question: Predict surface area and save results in csv format, only pcu topology.

Thought: I need to gather data on the surface area of materials with pcu topology and train a machine learning model to predict their surface area. Then, I can use the model to make predictions and save the results in a csv format.

Property: surface_area

Material: pcu*

Final Thought: The predicted surface area values for materials with pcu topology can be found in the saved csv file.

Question: {question}

**Supplementary Note S3.** Prompt for single data retrieval in predictor

You need to answer the question from the markdown table below

Markdown Table:

{table}

{information}

Question: {question}

Answer:

**Supplementary Note S4.** Prompt for planning in generator

---

Create a plan to generate material based on the following question.

Use the following format:

Question: the input question you must to answer

Thought: you should always think about what to do

Property: the property you can predict, must be one of [{model_names}]

Objective: you should decide what criteria you wan to generate by.

Search look-up table: plan to extract 100 material for the purpose from the look-up table where the property is pre-calculated.

Genetic algorithm: plan to create a new materials using the 100 extracted materials.

Final thought: get a final answer based on the structures you generate.

Begin!

Question: generate a material with a porosity of 0.5 and surface area of 120 m2/g

Thought: I need to generate a material with a porosity value of 0.5 and surface area of 120 m2/g.

Property: void_fraction, accessible_surface_area

Objective: near 0.5, near 120

Search look-up table: extract cif_id, void_fraction, and accessible_surface_area of 100 materials with porosity close to 0.5 and surface area near 120 m2/g from look-up tables.

Genetic algorithm: create a new material with a porosity close to 0.5 and surface area near 120 m2/g from 100 materials

Final Thought: Based on the generated CIF, find the material that is closest to a porosity of 0.5 and a surface area of 120 m2/g.

Question: generate a material with a highest band-gap

Thought: I need to generate a material with a highest band-gap.

Property: bandgap

---

Objective: max

Search look-up table: extract cif_id and bandgap of 100 materials with high band-gap value from look-up tables.

Genetic algorithm: generate 100 new materials with the highest band gap from the 100 materials.

Final Thought: Based on the generated CIF, find the material that has highest band-gap.

Question: {question}

**Supplementary Note S5.** Prompt for genetic algorithm in generator

"You should act as a generator to find the optimal material. A substance consists of a block1, block2, and must maintain the order. I will give you 100 parent materials. Based on these, you must answer as many new children as you expect to answer the question. The block1 and block2 used in the child must be the blocks used in the parent, and you must not create blocks that don't exist. You must generate children diverse. "The children must not be duplicates of existing parents or already created children. You output children only and nothing else.

Begin.

Question: {question}

Parent:

V12+T31, V24+T32, V7+T12

4 new Children:

V12+T12, V24+T31, V7+T31, V7+T32

Parent:

{parents}

100 new Children:

**Supplementary Note S6.** Prompt for planning in agent

You act like a material scientist answering a question. Answer the following questions as best you can. You have access to the following tools:

{tools}

You must not use tool `generator` when `Question` does not required to generate.

Use the following format:

Question: the input question you must answer

Thought: you should always think about what to do

Action: the action to take, should be one of [{tool_names}]

Action Input: the input to the action

Observation: the result of the action

... (this Thought/Action/Action Input/Observation can repeat N times)

Thought: I now know the final answer

Final Answer: the final answer to the original input question

Begin!

Question: What is the surface area and bandgap of ACOGEF?

Thought: I need to find the surface area of ACOGEF.

Action: search_csv

Action Input: "Search name ACOGEF and provide information on its surface area"

Observation: The surface area of material "ACOGEF" is 1138.35 m^2/g. Check to see if this answer can be you final answer, and if so, you should submit your final answer.

Thought: The search_csv tool provided the surface area of ACOGEF, but not the bandgap. I need to find the bandgap.

Action: search_csv

Action Input: "Search name ACOGEF and provide information on its bandgap"

Observation: The search_csv tool did not provide any information on the bandgap of ACOGEF. I need to find another way to obtain this information.

Thought: The search_csv tool provided the surface area of ACOGEF, but not the bandgap. I need to find the bandgap.

Action: predictor

Action Input: "Predict the bandgap of ACOGEF"

Observation: The bandgap of material "ACOGEF" is 3.41139 eV. Check to see if this answer can be you final answer, and if so, you should submit your final answer.

Thought: I now know the final answer

Final Answer: The bandgap and surface area of ACOGEF is 3.41149 eV and 1138.35 m^2/g.

Question: What is the highest bandgap MOF?

Thought: I need to find the MOF with the highest bandgap.

Action: search_csv

Action Input: "find for MOFs with highest bandgap and provide its name and bandgap"

Observation: There are no data in database

Thought: The search_csv tool did not provide any information on the bandgaps of MOFs. I need to find another way to obtain this information.

Action: predictor

Action Input: "predict the bandgaps of MOFs and find the name of MOF with highest bandgaps"

Observation: The highest bandgap MOF is ACOGEF.

Thought: I now know the final answer

Final Answer: The highest bandgap MOF is ACOGEF.

Question: {input}

Thought:{agent_scratchpad}

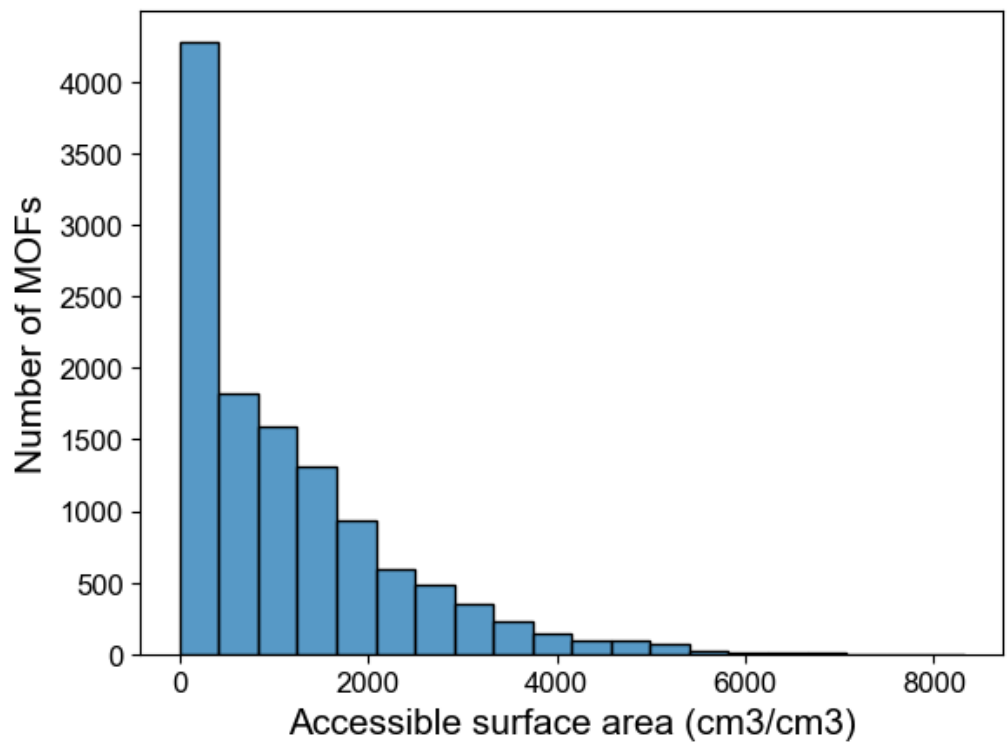

**Figure S1.** Histogram for accessible surface area of MOF in the CoREMOF database

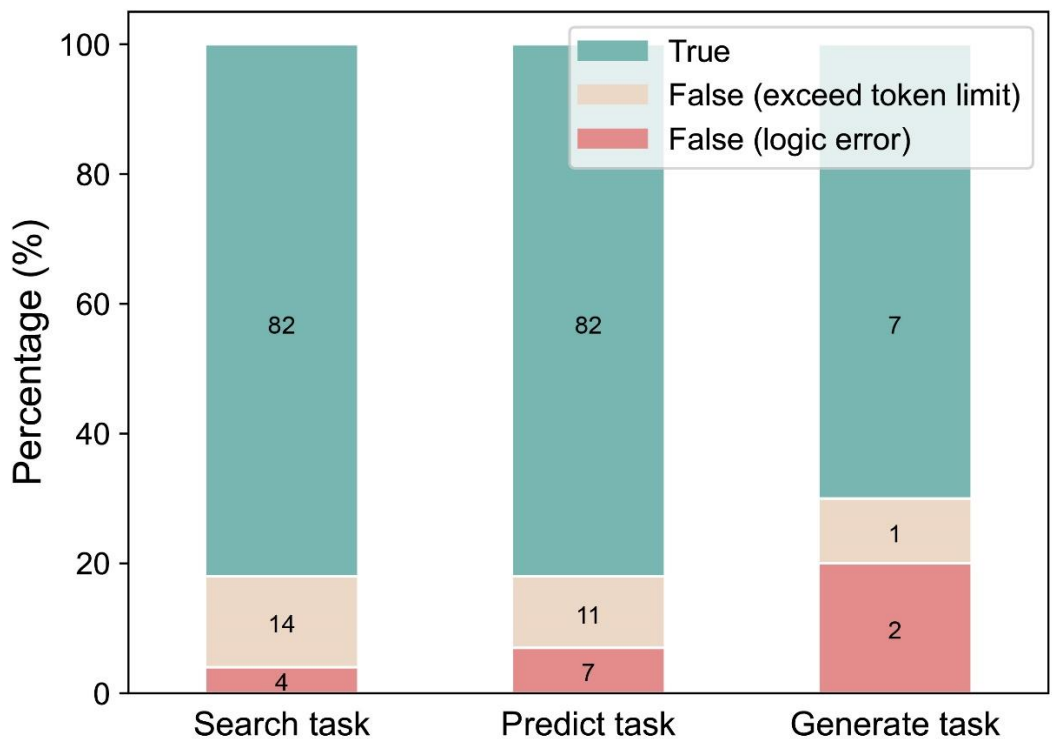

**Figure S2.** Depiction of accuracies for three tasks using gpt-3.5-turbo model - search, prediction, and generation. Accuracies were evaluated based on three labels: True, False (exceeding token limit), and False (logical error). The number in the bar represent the count of each label

**gpt-4**

```
########## Question ##########
How does the pore limiting diameter of YUSGID_clean compare with other materials?

> Entering new AgentExecutor chain...
I need to find the pore limiting diameter of YUSGID_clean and compare it with other
materials.
Action: search_csv
Action Input: "Search name YUSGID_clean and provide information on its pore limiting
diameter"

> Entering new TableSearcher chain...

[Table Searcher] Thought: To find the information for a specific name in the DataFrame,
we can use the loc function in pandas. This function allows us to access a group of rows
and columns by label(s) or a boolean array.
[Table Searcher] Input: print(df.loc[df['name'] == 'YUSGID']['Pore limiting diameter
(Å)'].to_markdown())
[Table Searcher] Observation:
|       | Pore limiting diameter (Å) |
|-------|:--------------------------:|
| 11739 |          3.71515           |
[Table Searcher] Final Thought: The 'Pore limiting diameter (Å)' for the material with the
name 'YUSGID' is 3.71515 Å.
> Finished chain.

Observation: The 'Pore limiting diameter (Å)' for 'YUSGID' is 3.71515 Å.
Thought:I have the pore limiting diameter for YUSGID_clean. Now I need to compare
this with the pore limiting diameters of other materials.
Action: search_csv
Action Input: "Provide information on the pore limiting diameters of all materials"

> Entering new TableSearcher chain...

[Table Searcher] Thought: To get information on the pore limiting diameters of all
materials, we can use the pandas function describe(), which provides descriptive
statistics of the DataFrame or Series.
[Table Searcher] Input: print(df['Pore limiting diameter (Å)'].describe().to_markdown())
[Table Searcher] Observation:
|       | Pore limiting diameter (Å) |
|-------|:--------------------------:|
| count |          12020             |
| mean  |          4.87759           |
| std   |          2.78185           |
| min   |          2.40008           |
| 25%   |          3.24153           |
| 50%   |          4.10336           |
| 75%   |          5.64609           |
| max   |          71.502            |
[Table Searcher] Final Thought: The describe() function in pandas provides a statistical
summary of the 'Pore limiting diameter (Å)' column in the DataFrame. This includes the
count, mean, standard deviation, minimum, 25th percentile, median (50th percentile),
75th percentile, and maximum.
> Finished chain.

Observation: The pore limiting diameters of all materials range from approximately 2.40Å
to 71.50Å, with an average diameter of approximately 4.88Å.
Thought:I now know the final answer
Final Answer: The pore limiting diameter of YUSGID_clean is 3.71515 Å, which is below
the average pore limiting diameter of approximately 4.88Å for all materials. The range of
pore limiting diameters for all materials is approximately 2.40Å to 71.50Å.
> Finished chain.

########## Output ##########
The pore limiting diameter of YUSGID_clean is 3.71515 Å, which is below the average
pore limiting diameter of approximately 4.88Å for all materials. The range of pore limiting
diameters for all materials is approximately 2.40Å to 71.50Å.
```

**gpt-3.5-turbo**

```
########## Question ##########
How does the pore limiting diameter of YUSGID_clean compare with other materials?

> Entering new AgentExecutor chain...
I need to compare the pore limiting diameter of YUSGID_clean with other materials.
Action: search_csv
Action Input: "Search name YUSGID_clean and provide information on its pore limiting
diameter"

> Entering new TableSearcher chain...

[Table Searcher] Thought: To search for a specific value in the "name" column, we
can use the pandas function loc[], which allows us to access a group of rows and columns by
label(s) or a boolean array. We can then extract the information on the pore limiting
diameter for the material with the name "YUSGID".
[Table Searcher] Input: print(df.loc[df['name'] == 'YUSGID', 'Pore limiting diameter
(Å)'].to_markdown())
[Table Searcher] Observation:
|       | Pore limiting diameter (Å) |
|-------|:--------------------------:|
| 11739 |          3.71515           |
[Table Searcher] Final Thought: The pore limiting diameter for the material with the
name "YUSGID" is 3.71515 Å.
> Finished chain.

Observation: The answer for question "Search name YUSGID and provide information
on its pore limiting diameter" is 3.71515 Å.
Thought:The search_csv tool provided the pore limiting diameter of YUSGID_clean. Now
I need to find the pore limiting diameter of other materials for comparison.
Action: search_csv
Action Input: "Search for other materials and provide information on their pore limiting
diameter"

> Entering new TableSearcher chain...

[Table Searcher] Thought: To search for other materials and provide information on their
pore limiting diameter, we can use the pandas function loc[] to filter the dataframe based
on a condition. We can then select the "name" and "Pore limiting diameter (Å)" columns
to display the information.
[Table Searcher] Input: print(df.loc[:, ['name', "Pore limiting diameter (Å)"]].to_mark-
down())
ValueError: The number of tokens has been exceeded.
```

Figure S3. ChatMOF's answer to the question "How does the pore limiting diameter of YUSGID_clean compare with other materials?" using gpt-4 and gpt-3.5-turbo. The answer is successfully derived for gpt-4 but raises a tokenError for gpt-3.5-turbo.

| |
|---|
| What is the largest cavity diameter of ASOVEL_clean? |
| What's the pore limiting diameter of XEXLUJ_clean? |
| How large is the free pore diameter for DIBQUC? |
| What's the density of ROLCEC19? |
| Can you tell me the accessible surface area of ALIBUT? |
| What is the non-accessible surface area of ARAHIM02? |
| What is the accessible volume fraction for EREGOY_clean? |
| Can you tell me about the accessible volume of UZANOZ? |
| What's the non-accessible volume for NAHFOU? |
| What type of metal is present in PITPEP_clean? |
| Does DUBKAO have an open metal site? |
| Can you tell me the type of open metal in FECYUJ? |
| Can you provide the top 5 materials with the highest accessible surface area? |
| Which materials have the highest non-accessible volume? |
| What's the largest cavity diameter in materials with a density greater than a 7.0? |
| Can you provide a list of materials with an open metal site? |
| What's the accessible volume fraction for materials with a density less than a 0.25? |
| How does the pore limiting diameter of YUSGID_clean compare with other materials? |
| What materials have the largest free pore diameter? |
| Is there an open metal site in NELVAC? |
| Can you provide a list of materials with the highest pore limiting diameter? |
| What materials have the largest accessible volume? |
| What's the non-accessible surface area for WAWGOQ_clean? |
| What is the type of open metal in YUNJIB? |
| Can you list the top 10 materials with the highest non-accessible surface area? |
| How high is the accessible surface area for JUKPAI_clean? |
| What's the largest cavity diameter in materials with a density less than a 0.8? |
| Can cg901114e_si_002 have an open metal site? |
| What materials have the largest accessible volume fraction? |
| What is the pore limiting diameter of LOLREL? |
| How does the accessible volume of OSIXEU compare with other materials? |
| Can you provide the top 10 materials with the highest density? |
| What's the largest free pore diameter in materials with a density less than a 0.8? |
| How high is the non-accessible volume for FIJDIM05? |
| What materials have the largest cavity diameter? |
| What is the metal type in FOTNIN? |
| How high is the accessible volume fraction for WOBCAT? |
| Can you list materials with a pore limiting diameter greater than a 6.0? |
| What is the density of materials with an accessible surface area above a 3000 m2/g? |
| What is the type of open metal in IYAFIX? |
| Can JILXOS have an open metal site? |
| What materials have the largest non-accessible surface area? |
| Can you list the top 5 materials with the highest largest free pore diameter? |
| What is the accessible volume for MOJJUR? |
| Can you provide the top 10 materials with the highest largest cavity diameter? |
| What is the accessible surface area for materials with a pore limiting diameter below a 2.0? |
| What is the non-accessible volume for WABTOK? |
| What materials have the largest pore limiting diameter? |
| What's the type of open metal in QOWRAV11? |
| Can PODVUB have an open metal site? |
| Can you list the top 10 materials with the highest accessible volume fraction? |
| What is the metal type in QULMIU? |
| How does the non-accessible surface area of YICTUZ compare with other materials? |
| Can you provide the top 10 materials with the highest accessible volume? |
| What's the pore limiting diameter in materials with a density greater than a 2.5? |
| How high is the largest free pore diameter for DUFKAS? |
| What materials have the largest density? |
| What is the largest cavity diameter for TONTIB_clean? |
| Can you list the top 5 materials with the highest non-accessible volume? |
| What is the accessible surface area for UXUPEJ? |
| What is the density of NATXIR? |
| How does the largest free pore diameter of LITDAV compare with other materials? |
| Can you provide the top 10 materials with the highest largest free pore diameter? |
| What's the pore limiting diameter for CSMNOX? |
| How high is the non-accessible surface area for MUZKAV_clean? |
| What materials have the largest accessible surface area? |
| What's the type of open metal in cm503311x_F-mil-53-Al-lp-mono? |
| Can TAHYUZ have an open metal site? |
| Can you list the top 5 materials with the highest density? |
| What is the metal type in YUBTUM? |
| How does the accessible surface area of TATFOL compare with other materials? |
| Can you provide the top 5 materials with the highest accessible surface area? |
| What's the largest cavity diameter for NELVAC? |
| How high is the accessible volume for EPUJIJ_clean? |
| What materials have the largest non-accessible surface area? |
| What's the type of open metal in WAWGOQ_clean? |
| Can YUNJIB have an open metal site? |
| Can you list the top 5 materials with the highest accessible volume fraction? |
| What is the metal type in JUKPAI_clean? |
| How does the non-accessible surface area of HUZFIS compare with other materials? |
| Can you provide the top 5 materials with the highest largest cavity diameter? |
| What's the pore limiting diameter for XIPTAS_clean? |
| How high is the largest free pore diameter for LOLREL? |
| What materials have the largest density? |
| What is the largest cavity diameter for YAVVUP? |
| Can you list the top 5 materials with the highest non-accessible volume? |
| What is the accessible surface area for FIJDIM05? |
| What is the density of TUFTAR? |
| How does the largest free pore diameter of FOTNIN compare with other materials? |
| Can you provide the top 5 materials with the highest largest free pore diameter? |
| What's the pore limiting diameter for LAXGOH02? |
| How high is the non-accessible surface area for GEDQOX? |
| What materials have the largest accessible surface area? |
| What's the type of open metal in JILXOS? |
| Can VAHSON have an open metal site? |
| Can you list the top 5 materials with the highest density? |
| What is the metal type in MOJJUR? |
| How does the accessible surface area of REHVEH_clean compare with other materials? |
| Can you provide the top 5 materials with the highest accessible surface area? |
| What's the largest cavity diameter for NELVAC? |

**Table S1.** 100 questions in search task.

| |
|---|
| At 77 K, what's the hydrogen uptake for ZAXQOG_clean at 100bar? |
| Under standard temperature conditions of 298K and dilute system, can you list the materials with the highest O2 diffusivity? |
| Is XEGKUR_clean thermally stable? |
| What's the thermal stability of QINSUD_clean? |
| Which materials have a bandgap between 1 and 2 eV? |
| At room temperature (298K), what's the CO2 Henry coefficient for XEGKUR_clean? |
| Can you provide the top 10 materials with the highest N2 diffusivity at 298K and infinity dilute system? |
| Is the solvent removal stability of XEGKUR_clean strong? |
| Can you list 5 materials with a bandgap near 1.6 eV when evaluated at 298K and 1bar? |
| Under 77K and infinity dilute conditions, what's the hydrogen diffusivity for NISPEL_charged? |
| What materials show the best performance in hydrogen uptake at 100bar considering a temperature of 77 K? |
| How high is the O2 uptake of LOQLIN_SL at 298K and normal pressure? |
| Can XAVDUV_clean stably store hydrogen under 100bar pressure when observed at 77K? |
| At 298K and standard atmospheric pressure, how high is the N2 uptake for UXABUR_clean? |
| What's the O2 diffusivity for OCUVUF_clean under conditions of 298K and dilute system? |
| Is XAHWAG_clean stable when the solvent is removed? |
| What are the top 5 materials with the highest CO2 Henry coefficient at room temperature? |
| At 298K and dilute system, how high is the N2 diffusivity for XEGKUR_clean? |
| What materials are known for their high O2 uptake at a temperature of 298K and 1 bar pressure? |
| How high is the bandgap for XEGKUR_clean? |
| What materials have the highest thermal stability? |
| Can you show materials with a bandgap greater than 3.0 eV? |
| At 298K and standard atmospheric pressure, how much O2 can XEGKUR_clean take up? |
| What is the CO2 Henry coefficient for materials with bandgap less than 1.5 eV at 298K? |
| At 77K, how does the hydrogen uptake of OCUVUF_clean compare with other materials at 100bar? |
| Can you list materials with hydrogen diffusivity greater than 0.1 cm2/s at 77K and dilute system? |
| What is the N2 diffusivity in XEGKUR_clean under 298K and dilute conditions? |
| What is the thermal stability of materials with bandgap above 2.5 eV? |
| Which materials demonstrate a high level of O2 diffusivity at 298K and dilute condition? |
| How does the N2 uptake of XEGKUR_clean compare with others at 298K and 1 bar? |
| Can QINSUD_clean remove solvents stably? |
| How high is the thermal stability for NISPEL_charged? |
| At 298K and 1 bar pressure, what is the O2 uptake of MIFROK_clean? |
| Under standard conditions of 298K and 1 bar, which materials have the highest N2 uptake? |
| Can you list materials that show solvent removal stability? |
| How does the hydrogen diffusivity for QINSUD_clean compare with others at 77K and dilute system? |
| Under conditions of 298K, what is the CO2 Henry coefficient for WOYJOL_clean? |
| What materials have the best N2 diffusivity when measured at 298K and dilute system? |

Is the bandgap of XEGKUR_clean within the range 0.5-1.5 for photovoltaic applications?

At standard conditions of 77K and 100 bar, which materials have the highest hydrogen uptake?

How high is the bandgap for XEGKUR_clean when measured at 298K and under a pressure of 1 bar?

Can you provide a list of materials with high thermal stability?

At 77K and 100bar pressure, what is the hydrogen uptake of OCUVUF_clean?

What is the solvent removal stability of XAHWAG_clean?

Can GUCJAQ_clean diffusively handle N2 when evaluated at 298K and dilute system?

What materials show high O2 uptake capabilities under conditions of 298K and 1bar?

At 298K and dilute system, what is the diffusivity of O2 in XEGKUR_clean?

How much N2 can DUVNIS01_charged take up at 298K and 1 bar?

Under conditions of 298K, which materials have a high CO2 Henry coefficient?

How much hydrogen can XEGKUR_clean take up at 100bar pressure when observed at 77 K?

Under 298K and dilute conditions, how high is the O2 diffusivity for DUVNIS01_charged?

At standard temperature (298K) and 1 bar pressure, can IZEHAX_clean hold a high volume of N2?

At 298K and 1bar, how stable is JALCAD_clean when the solvent is removed?

What is the thermal stability of COKMUM_clean?

Can AYOYOE_clean handle a high volume of O2 when tested at 298K and 1 bar?

At 298K and dilute system, what is the diffusivity of N2 in IHAJIL_clean?

| |
|---|
| What materials show a high level of N2 diffusivity under conditions of 298K and 1 bar pressure? |
| Is the bandgap of XEGKUR_clean higher or lower than 1.5 eV? |
| At a temperature of 298K and a pressure of 1 bar, how does the O2 uptake of DUVNIS01_charged compare with others? |
| Under standard ambient temperature (298K), what is the CO2 Henry coefficient for XEGKUR_clean? |
| At 77K and 100 bar, can DUVNIS01_charged take up a significant volume of hydrogen? |
| Considering conditions of 298K and 1 bar, how much N2 can IZEHAX_clean hold? |
| At 77 K and dilute system, what is the hydrogen diffusivity for JALCAD_clean? |
| Can COKMUM_clean remain stable when the solvent is removed? |
| Which materials exhibit the best O2 diffusivity at 298K and dilute system? |
| How high is the bandgap for XEGKUR_clean? |
| What is the thermal stability of DUVNIS01_charged? |
| Under a temperature of 298K and pressure of 1 bar, how high is the O2 uptake for IZEHAX_clean? |
| At 298K and standard pressure, what materials have the highest N2 uptake? |
| Can you list the top materials that demonstrate thermal stability? |
| Given the conditions of 298K and dilute condition, what is the N2 diffusivity for QINSUD_clean? |
| What materials are thermally stable? |
| For a metal-organic framework at 298K and 1 bar, what is the CO2 Henry coefficient for KUGQIN_clean? |
| Can you provide the materials with the highest O2 diffusivity when measured at 298K and dilute system? |
| What materials have a high bandgap? |
| Considering 77K and 100bar conditions, how high is the hydrogen uptake of XEGKUR_clean? |
| Under ambient conditions of 298K and 1 bar, what is the N2 uptake for DUVNIS01_charged? |
| Can OCUVUF_clean remove solvents without compromising its stability? |
| At a standardized temperature of 77K and dilute system, how does XAHWAG_clean's hydrogen diffusivity measure? |
| What is the O2 uptake of GUCJAQ_clean at the typical conditions of 298K and 1 bar? |
| Can GAJTUI_clean take up a significant volume of N2 under conditions of 298K and 1 bar? |
| How stable is JEDJUY_clean at elevated temperatures? |
| At standard ambient conditions, which materials exhibit a high CO2 Henry coefficient? |
| Under the set conditions of 77K and dilute, how high is the hydrogen diffusivity of XEGKUR_clean? |
| What is the solvent removal stability of DUVNIS01_charged? |
| At 298K and a infinity dilute condition, how does OCUVUF_clean's N2 diffusivity compare to other materials? |
| Which materials show a pronounced hydrogen uptake at 77K and 100 bar? |
| Can XEGKUR_clean stably take up a large volume of O2 when examined at 298K and 1 bar? |
| What is the thermal stability of DUVNIS01_charged? |
| At standard conditions (298K and 1 bar), which materials are known for high O2 uptake? |
| Can XEGKUR_clean stably remove solvents? |
| How high is the bandgap of DUVNIS01_charged? |
| Which materials exhibit the highest N2 diffusivity at a temperature of 298K and infinity dilute condition? |
| Under conditions of 298K and 1 bar, how elevated is the N2 uptake for XEGKUR_clean? |
| At the standardized 298K and infinity dilute system, what is the O2 diffusivity of DUVNIS01_charged? |
| Under standard conditions, how high is the CO2 Henry coefficient for IZEHAX_clean? |
| At 77K and 100 bar pressure, can JALCAD_clean stably accommodate a large volume of hydrogen? |
| Under the given conditions of 77K and dilute system, which materials have the highest hydrogen diffusivity? |
| Is XEGKUR_clean thermally stable? |
| Can you provide a list of materials with a bandgap exceeding 2 eV? |

**Table S2.** 100 questions in prediction task.

| |
|---|
| Can you generate the structures with the largest surface area? |
| I'm looking for structures with a hydrogen uptake of about 500 cm3/cm3 at 100bar and 77K, can you generate those? |
| Can you generate structures with a density close to 0.4? |
| What structures can you generate with a smallest accessible pore volume? |
| I'm interested in structures with a low O2 uptake at 298K and 1bar, can you generate these? |
| I'm curious about structures where accessible volume fraction is less than 0.4, can the model generate these? |
| Can the model generate structures where the largest cavity diameter is in the bottom quartile? |
| I am curious about structures where the thermal stability is the highest among all, can you generate those? |
| What structures can be generated with a N2 uptake in the 3rd quartile range? |
| Can you create a MOF with a void fraction greater than 0.5 that has the highest CO2 Henry coefficient at 298 K? |

**Table S3.** 10 questions in generation task